\documentclass[runningheads]{llncs}
\usepackage{graphicx}
\usepackage{hyperref}
\usepackage{amsmath}
\usepackage{amsfonts}
\usepackage{mathrsfs}
\usepackage{color}
\newcommand{\vektor}[1]{\vec{\mathbf{#1}}}
\newcommand{\arr}[1]{\mathbf{#1}}

\begin{document}

%
\title{Addressing multiple metrics of group fairness in data-driven decision making}
%
%
\author{Marius Miron \inst{1} \and
Song\"ul Tolan\inst{1} \and
Emilia G\'{o}mez \inst{1,2} \and
Carlos Castillo \inst{2}}
\authorrunning{Miron et al.}
%
\institute{Joint Research Centre, European Commission\\
\email{\{name.lastname\}@ec.europa.eu}\\
 \and
Universitat Pompeu Fabra, Barcelona\\
\email{\{name.lastname\}@upf.edu}}
\maketitle              
\begin{abstract}
The Fairness, Accountability, and Transparency in Machine Learning (FAT-ML) literature proposes a varied set of \emph{group fairness} metrics to measure discrimination against socio-demographic groups that are characterized by a \emph{protected feature}, such as gender or race.
Such a system can be deemed as either \emph{fair} or \emph{unfair} depending on the choice of the metric. Several metrics have been proposed, some of them incompatible with each other.
%
We present here a framework to navigate the tensions between various group-wise metrics and to study fairness in data-driven decision making without the constraint of choosing a single metric.
We do so empirically, by observing that several of these metrics cluster together in two or three main clusters for the same groups and machine learning methods. 
%
In addition, we propose a robust way to visualize multidimensional fairness in two dimensions through a
Principal Component Analysis (PCA) of the group fairness metrics. Experimental results on multiple datasets show that the PCA decomposition explains the variance between the metrics with one to three components.
\end{abstract}
\newpage


\section{Introduction}

Machine Learning (ML) systems reduce uncertainty in decision-making by predicting relevant outcomes based on algorithmically detected patterns in data.
%
%
However, a growing literature has uncovered \emph{algorithmic discrimination} in sensitive contexts and described fairness-aware ML algorithms~\cite{Barocas2016,hajian2016algorithmic}.
A number of statistical formalizations of a value-driven concept such as fairness, constructed for the purpose of using them in data-driven algorithms, have led to a long and confusing list of criteria and related metrics~\cite{narayanan2018translation}.
These criteria suffer from incompatibilities and trade-offs between fairness and other objectives, such as accuracy.
To better understand the relation between conflicting criteria we study how a varied set of group-wise classification metrics are related across multiple datasets. For instance, by determining which metric yields more disparity between groups, one can better explain the type of discrimination in a machine learning model and focus on optimizing that specific metric. 

There are two main families of algorithmic fairness concepts: individual fairness~\cite{dwork2012fairness} and group fairness.
The latter derives from a concept of non-discrimination on the basis of membership to a protected group; the majority of the existing literature on Fairness, Accountability, and Transparency in Machine Learning (FAT-ML) refers to it and it is the focus of this paper.
%
%
A protected group is a group distinguished by a protected feature~\cite{pedreschi2012study}, where protected features are usually categories from a given legal context, such as gender or race in anti-discrimination legislation.\footnote{See e.g. Article 21 of the EU Charter of Fundamental Rights: \href{https://fra.europa.eu/en/charterpedia/article/21-non-discrimination}{https://fra.europa.eu/en/charterpedia/article/21-non-discrimination}}

In the case of automatic classification algorithms, group fairness is the absence of group discrimination, and group discrimination is evidenced by imbalances with respect to classification metrics across protected and non-protected groups.
These group-wise metrics are usually defined in terms of predicted risk scores, predicted outcomes, actual outcomes, or a combination of them.

Fulfilling specific fairness criteria in a machine learning algorithm is usually done through either processing training data, modifying the way the algorithm works, or modifying the output of the algorithm (pre-, in-, and post-processing \cite{Barocas_mimeo}).
In most cases, the unifying idea is that the objective of minimizing a loss-function is constrained to the fulfillment of fairness criteria.
Clearly, imposing many fairness constraints will make finding an optimum impossible, and in fact, ``impossibility theorems'' prove that multiple fairness criteria are incompatible under fairly weak assumptions  \cite{kleinberg2016inherent,Chouldechova2017,berk2017convex}.
However, since different fairness notions lead to different fairness criteria, and not all notions can be fulfilled with just one criterion, algorithm developers are left with the dilemma of deciding between different value-concepts when trying to implement an appropriate fairness metric into the algorithm \cite{komiyama2018nonconvex}.

In this paper, we develop a framework that helps better deal with this problem by shifting the decision margin: instead of choosing the appropriate fairness metric in a given decision setting, we assess disparate impact by considering a higher number of group classification metrics. Our contributions are listed below:

\textbf{Clustering of metrics }
We start by showing empirically, that many group metrics are highly correlated and can be clustered into two or three groups, which is in line with findings by Friedler et al. \cite{friedler2019comparative}.
%
%
Moreover, a Principal Component Analysis (PCA) uncovers how a multidimensional vector of group metrics can in practice be reduced to two main axes. This also allows for the ranking of different ML algorithms along the line of these axes.

\textbf{Fairness visualization }
In addition to the empirical analysis, our contributions go in the direction of fairness visualization. The clustering displays a multitude of fairness-related factors and their correlations in a single graph allowing researchers to focus on a smaller set of metrics that are orthogonal to each other.
%
%
Furthermore, in order to better visualize the disparities, we align and center the PCA decomposition of various matrices corresponding to different ML models.   

The remainder of this paper is structured as follows. In Section \ref{sec:relation} we discuss the relation with the previous work. In Section \ref{sec:methodology} we present the methodology including the problem definition, a method to visualize multiple group fairness metrics based on clustering, a method to evaluate correlations of these metrics between various datasets, and a method to compare ML models in terms of fairness using PCA. We present experiments using the proposed framework in Section \ref{sec:experiments}. In Section \ref{sec:conclusions} we present the conclusions. 

\section{Relation to previous work}\label{sec:relation}

The FAT-ML literature is voluminous, multidisciplinary, and rapidly evolving. Reviews on fairness criteria in decision making are provided by Romei and Ruggieri~\cite{romei2014multidisciplinary}, \v{Z}liobait\.{e}~\cite{zliobaite2017measuring}, as well as Barocas and Selbst\cite{Barocas2016} who elaborate on mechanisms to address biased data and algorithmic unfairness.
There are different methods to ensure that fairness criteria are satisfied in classification algorithms \cite{kamiran2012data,calmon2017optimized,Hardt2016,pleiss2017fairness,Zafar2017fairness,agarwal2018reductions,komiyama2018nonconvex}. 
%
In this research we do not propose a fair ML algorithm. Our goal is to study the disparity that occurs when deploying general-purpose ML algorithms. Specifically, we propose a framework to analyze for fairness, similarly to other frameworks which test software for discrimination, such as Themis \cite{galhotra2017fairness}, Aequitas \cite{saleiro2018aequitas}, and BlackBoxAuditing \cite{adler2018auditing}. Following the recommendations in \cite{friedler2019comparative} and in contrast to Themis and BlackBoxAuditing, we do not explore other metrics in the realms of causal discrimination or indirect influence.\footnote{Our framework has the possibility of exploring feature importance, however this is not the goal of this paper.} 
Similar to Aequitas which provides a map to navigate between different group-fairness metrics according to the type of problem, we aim at giving a bigger picture on how these metrics are related and we plot the disparity between various protected groups in a lower dimensional space. 

The choice of an appropriate fairness measure from a long list of potential ones can be very complex, as it depends on the respective policy context and the stakeholders involved~\cite{narayanan2018translation}.
For instance, \cite{liu2018delayed,menon2018cost} restrict their analysis on two fairness metrics: true positive rates and predictive prevalence. Another fairness-enhancing method \cite{Zafar2017fairness} optimizes for false positive rates and false negative rates, a constraint which is also discussed in \cite{berk2017convex}.
Furthermore, there are impossibility theorems \cite{kleinberg2016inherent,Chouldechova2017} that mathematically prove the impossibility of reconciling different fairness notions if the prevalence of the outcome, i.e., the ``base rate'' differs across different protected groups.
There is a tension between fairness criteria and optimal accuracy \cite{Corbett2017,Zafar2017fairness,berk2017convex,menon2018cost}. To that extent, in this paper we study the conflict between group-wise accuracy metrics and fairness related metrics. Particularly, we study how the group-wise binary classification metrics relate to each other using clustering and correlation. 

%

Another approach to addressing the complexity in fair machine learning is to simplify the long list of fairness criteria before addressing its tensions.
Our approach relates most to the one taken by
Friedler et al.~ \cite{friedler2019comparative} who show that most fairness measures are highly correlated with one another. 
As a further extension of their work, we use clustering to have a more in depth view on how the measures are related to each other. 
%
%
%
After performing a PCA of all tested fairness measures, our approach goes one step further by extending the framework to the comparison of algorithms.


\section{Methodology}\label{sec:methodology}

\subsection{Problem definition}\label{ssec:problemdef}
A dataset contains a set of features $\vektor{x}=[x_0\dots x_F]$, including a set of protected features $\vektor{z}=[z_0\dots z_P]$, with $\vektor{z} \in \vektor{x}$, and a set of associated binary predictions $\vektor{y}=\{0,1\}$. A binary decision making system takes as input the features $\vektor{x}$ and solves a binary classification task with an output $\vektor{y}$ where the binary labels have different meanings depending on the task, e.g., not re-offended/re-offended, bad/good credit score, not receiving/receiving a benefit. Depending on the impact on the human subjects, a decision is assistive, as in the case of credit scoring (should/should not receive a loan), or punitive, as in the case of recidivism\footnote{Recidivism is defined as the act of a person committing a crime after they have been convicted of an earlier crime \cite{brennan2013emergence}.} prediction (recidivist/non-recidivist) in criminal justice.
A machine learning model solving this binary classification task yields a set of predictions $\hat{\vektor{y}}=\{0,1\}$.
The performance of the machine learning model is usually measured on test data comprising $N$ pairwise observations $(\arr{X}^{N\times F},\hat{\arr{Y}}^N)$ and their associated ground-truth binary labels $\arr{Y}^N$.

\textbf{Group metrics:} Let $\vektor{m}(j)=[m_0,\dots,m_J]$ be a set of metrics used to report the performance of a given machine learning model on the test set such as the ones defined in Section \ref{ssec:metrics} (accuracy, false positive rate etc.). 
In the literature, group fairness is defined for a particular metric $\vektor{m}(j)$ in relation to a protected feature $p$ (e.g., gender).
Hence, given a protected feature $p$ (e.g., gender) from $\vektor{z}$ and the associated groups $z_p=\{z_p(1),\dots,z_p(G)\}$ (e.g.,\{ Men, Women \}), we compute the group-wise metrics $\vektor{m_g}$ for each $g=[1,\dots,G]$ (i.e., given gender as the protected feature, we compute the metrics for the groups Men and Women).  
An outcome is considered fair with respect to the metric $\vektor{m}(j)$ and two groups  $(g,h)$ if $\vektor{m}_{g}(j)/\vektor{m}_{h}(j)=1$.
%
%

In this paper we aim at developing a method to evaluate group fairness that encompasses several metrics in $\vektor{m}_{g}$ rather than relying on a single metric $\vektor{m}_{g}(j)$. To that extent, we want to compare $l=\{1,\dots,L\}$  ML models for all groups $g=\{0,\dots,G\}$ of a given protected feature $p$ across all the metrics $\vektor{m}_{g}$.

\subsection{Clustering of metrics}\label{ssec:clusterigmet}
Here we aim at discovering relations between different metrics, at finding out which produces more discrimination, and at determining which groups are more discriminated across multiple features. 
We achieve this by clustering the metrics vectors $\vektor{m}_{g,l,p}$ computed for each group $g$ and machine learning method $l$ at each protected feature $p$. 
For each protected feature $p$ we form the matrix $\arr{M}_p$ with the vectors $\vektor{m}_{g,l,p}$ on each line: $\arr{M}_p=(\vektor{m}_{1,1,p} \dots \vektor{m}_{G,L,p})$. The matrix $\arr{M}_p$ has the size $(I,J)$ where $I=G\cdot L$.

\textbf{Clustering the columns: }We cluster the columns to analyze the relation between metrics. We compute the pairwise correlation between the columns of this matrix $\vektor{d}^I_p(j,j\prime)=\rho(\arr{M}_p(j),\arr{M}_p(j\prime))$ where $j,j\prime =\{1,\dots ,J\}$.
Then, we perform a hierarchical clustering of the metrics using $\vektor{d}^I_p(j,j\prime)$.
Clusters are merged and created using the un-weighted pair grouping method~\cite{day1984efficient}.

\textbf{Clustering the rows: }We cluster the rows, ML models and groups, to discover how far are different groups from each other for each ML model. This clustering involves computing the distance vector $\vektor{d}^J_p(i,i\prime)=\rho(\arr{M}_p(i),\arr{M}_p(i\prime))$ where $i,i\prime =\{1,\dots ,I\}$.

In order to see if the clusters yielded by the distance vectors $\vektor{d}^J_p$ are similar across different datasets and protected features, we compute the correlation between the distance vectors.
A high correlation means that the metrics produce a similar clustering across different datasets. 

\subsection{PCA decomposition on the columns and visualization}\label{ssec:pcamet}

Due to the high number of metrics $J$, it is cumbersome to visualize the distance between $G$ groups and $L$ ML models using the matrix $\arr{M}_p$ of dimensions $(I,J)$. It is therefore desirable to reduce the number of $J$ metrics by projecting the columns of the matrix $\arr{M}_p$ to a lower dimension. 

Because we need to compare distances between a reference group, usually the largest group, and the other groups across different ML methods, a straight-forward PCA decomposition of the matrix $\arr{M}_p$ produces a 2D or 3D scatter plot of all data points in the matrix which makes it difficult to compare between various ML models.
In order to have a fair comparison between ML models we need to plot the disparity between the reference group and the other groups for each ML model separately. 
Thus, we need to form metrics matrices for each ML model $\arr{M}_{p,l}$ and we obtain the basis vectors for $l=1$. The PCA is then obtained by multiplying each $\arr{M}_{p,l}$ with the basis vector. For a better visualization of disparity between groups, we align and overlay these plots with the axes centered on the reference group.
%
%

For each protected feature $p$ we form a matrix $\arr{M}_{p,l}=(\vektor{m}_{1,l,p},\dots ,\vektor{m}_{G,l,p})$ which contains solely the metrics for a ML model $l$. 
Then, for $l=1$ we apply a PCA decomposition to the matrix $\arr{M}_{p,1}$ obtaining the eigenvectors $\arr{E}_{p,1}$. Then, the PCA decomposition is computed as $\hat{\arr{M}}_{p,l}=\arr{E}_{p,1}\arr{M}_{p,l}$ for all $l=(1\dots L)$. 
Let $R$ be the index of the largest group in $(1\dots G)$. Then, we align the matrix $\hat{\arr{M}}_{p,l}$ with respect to $R$ by subtracting the vector $\vektor{m}_{R,l,p}$ from each row of the matrix. 

%
%

%

%
%
%


\section{Experiments}\label{sec:experiments}

\subsection{Datasets}\label{ssec:datasets}

\subsubsection{YouthCAT: Recidivism in juvenile justice in Catalonia.}\footnote{Recidivism is defined as the act of a person committing a crime after they have been convicted of an earlier crime \cite{brennan2013emergence}.}
This dataset contains data on juvenile recidivism in Catalonia for offenders aged 12-17 years (N=4,753)\footnote{Provided by the Centre for Legal Studies and Specialised Training \cite{cejfe_2017_savry}, available online \href{http://cejfe.gencat.cat/en/recerca/opendata/jjuvenil/reincidencia-justicia-menors/index.html}{http://cejfe.gencat.cat/en/recerca/opendata/jjuvenil/reincidencia-justicia-menors/index.html}}
The crimes have been committed between 2002 and 2010 and all sentences were finished by 2010. The recidivism was reported in 2013 and 2015. The dataset contains demographic (age, foreign status, nationality: Spanish, European, Latin American, Maghrebi, Other) and criminal history features (number of crimes, type of crime, sentence).
This dataset has been used in the following study on algorithmic fairness: \cite{Miron2019}.

\subsubsection{COMPAS: Recidivism risk score in Broward County.}
Correctional Offender Management Profiling for Alternative Sanctions (COMPAS) is a risk assessment tool developed by Northpointe which assesses a criminal defendant’s likelihood to re-offend \footnote{Provided by ProPublica, available online \href{https://github.com/propublica/compas-analysis/}{https://github.com/propublica/compas-analysis/}.}. The database contains: criminal history, jail and prison time, demographics and COMPAS risk scores for defendants from Broward County from 2013 and 2014. The data comprise demographic features (age, gender, race) and criminal history features (count of prior crimes, type of crime).
This dataset has been used in the following studies on algorithmic fairness: \cite{Chouldechova2017,Corbett2017}.

\subsubsection{Statlog: German Credit Dataset.}
This dataset has been used to classify people (N=1,000) described by a set of attributes as good or bad credit risks \footnote{Provided by Hamburg University, available online \href{https://archive.ics.uci.edu/ml/datasets/statlog+(german+credit+data)}{https://archive.ics.uci.edu/ml/datasets/statlog+(german+credit+data)}.}. As attributes, the dataset contains demographic (age, foreign status, gender and marital status) and qualitative features (status of account, savings, credit history, purpose).
This dataset has been used in the following studies on algorithmic fairness:  \cite{zemel2013learning,fish2016confidence}

\subsubsection{Credit: Default of Taiwanese Credit Card Clients.}
This dataset \cite{yeh2009comparisons} has been used to detect default payments in Taiwan (N=30,000) \footnote{Provided by Chung Hua University, Taiwan, available online \href{http://archive.ics.uci.edu/ml/datasets/default+of+credit+card+clients}{http://archive.ics.uci.edu/ml/datasets/default+of+credit+card+clients}.}. These data include demographic features such as age, gender, marital status, education, as well as credit history data including amount of given credit, amount of bill payment, and history of past payment.
This dataset has been used in the following studies on algorithmic fairness: \cite{berk2017convex,lipton2018does}\\

These datasets differ from other datasets used in binary classification due to the fact that they are cases of decision making processes which can be automated using machine learning. According to the classification in Section \ref{ssec:problemdef} the decisions for case of credit score (Statlog) are assistive, and in the case of recidivism prediction (COMPAS, YouthCAT) and credit card default (Credit) are punitive.

Due to the impact on human subjects, the decisions have an important ethical dimension \cite{Barocas2016} and are often addressed from the point of view of fairness in decision making as indicated by their use in the FAT-ML literature. In this case we are interested in group fairness, whether the machine learning decisions are biased towards a particular category of people. The protected features  are \textit{gender} for all four datasets, \textit{foreigner} status for YouthCAT and Statlog, and \textit{national group} and \textit{race} for YouthCAT and COMPAS.

\subsection{Machine Learning Methods}

Each of the datasets described in \ref{ssec:datasets} proposes a decision making problem which can be modeled as a binary classification task: predicting whether someone will recividate or not (COMPAS, YouthCAT), predicting if someone will default or not (Credit), and predicting whether a person is a good creditor (Statlog).

We test a number of machine learning algorithms for supervised learning: logistic regression (\textit{\textit{logit}}), multi-layer perceptron (\textit{mlp}), support vector machine with a linear (\textit{lsvm})  K-nearest neighbors (\textit{knn}), random forest (\textit{rf}), decision trees (\textit{tree}), and naive bayes (\textit{nb}) \cite{robert2014machine}.

To account for overfitting we use cross-validation to split the data between training and testing. 
In each split, the validation data is chosen from the training set, with $10$\% random elements kept for validation. The validation set is used to tune the ML models hyper-parameters and to pick the binarization threshold for the prediction of the ML models.

Fairness-aware machine learning aims at fixing the disparities in ML algorithms with respect to a single metric. Due to the fact that these methods operate at different steps (pre-, post-, or during training) and can optimize with respect to different metrics, we do not attempt a detailed comparison between these methods in this paper.

\subsection{Metrics}\label{ssec:metrics}
Here we aim at computing a set of performance metrics from which we derive a set of group-wise metrics corresponding to the protected features and the groups in each dataset. 
There are various metrics for evaluating a ML method, some of these are application-dependent. For instance, a ML system in a criminal justice context might be evaluated differently from one in e-commerce.
Furthermore, the meaning of a metric changes when the decision making intervention is assistive ($\vektor{y}=1$ means a good creditor in Statlog) or punitive ($\vektor{y}=1$ means a recidivist in COMPAS).

\subsubsection{Performance metrics.}
The ML predictions $\hat{\arr{Y}}$ and their associated ground-truth binary labels $\arr{Y}$ determine four numbers: true positives $TP$ (correct positive assignments), true negatives $TN$ (correct negative assignments), false positives $FP$ (incorrect positive assignments), and false negatives $FN$ (incorrect negative assignments).
From $TP,FP,TN,FN$ we can calculate various metrics including the True Positive Rate, $TPR=TP/(TP+FN)$, with the complement False Negative Rate, $FNR=FN/(TP+FN)$, the True Negative Rate, $TNR=TN/(TN+FP)$, with the complement False Positive Rate, $FPR=FP/(TN+FP)$.
Furthermore, we can compute Positive Predictive Value, $PPV=TP/(TP+FP)$, with complement the False Discovery Rate, $FDR=TN/(TN+FN)$; and Negative Predictive Value, $NPV=TN/(TN+FN)$, with complement the False Omission Rate, $FOR=FN/(TN+FN)$.
Furthermore, we compute metrics which depend on the prevalence: Predicted Prevalence $PPREV=(TP+FP)/N_g$ and Predicted Positive Rate $PPR=(TP+FP)/N$, where $N$ is the total number of people in the dataset and $N_g$ is the total number of people in the data which are part of a group $g$. In this case, $PPREV$ solely makes sense as a group-wise metric.

We compute Accuracy, which measures how well a model correctly detects or excludes a condition $(TP+TN)/(TP+TN+FP+FN)$, and Balanced Accuracy $BA=(TPR+ TNR)/2$.

If ML predictions $\hat{\arr{Y}}$ are probabilistic then these metrics can be obtained at different classification thresholds which are applied to the output probability. To measure predictive performance we use the area under the ROC curve (AUC) which trades-off specificity (false positive rate) and sensitivity (true positive rate) for all the thresholds $t=(0,1)$.

\subsubsection{Computing a classification threshold.}
To maximize the performance of the models we choose the threshold value which maximizes balanced accuracy on the validation set, defined as
$BA(t)=(TPR(t) + TNR(t))/2$,
where $t=(0,1)$ is the varying threshold, $TPR$ is the true positive rate, and $TNR$ is the true negative rate. The best threshold $t_{max}$ is obtained for $max(BA)$ on the validation set. Here we report $BA$ as $max(BA)$ on the test set.

\subsubsection{Group-wise metrics.}
As described in Section \ref{ssec:problemdef} the classification metrics $\vektor{m}=\{AUC, A, BA, FPR, TPR, FNR, TNR, PPV, NPV, FDR, FOR, PPR,$ \linebreak$ PPREV\}$ are computed group-wise, to evaluate the fairness of ML models~\cite{narayanan2018translation}. Thus, we calculate $\vektor{m}_{g,l,p}$ for each experiment, for each protected feature $p$ and the corresponding groups $g$, and for all the ML models $l$.

\subsection{Experimental setup}

\subsubsection{Data encoding.}
Data is encoded numerically. 
Numerical values are normalized to have a mean of $0$ and standard deviation of $1$.
Categorical features are encoded as binary if they have two possible values, or using one-hot encoding when having multiple categories.
Ordinal features (e.g., low, medium, high) are also encoded numerically.

\subsubsection{Parameters and model selection.}
We perform k-fold cross validation with $k=10$. Each fold is replicated $10$ times with a different random seed that controls the random split between training, validation, and testing sets.

While for some ML classifiers (\textit{nb}, \textit{logit}, \textit{mlp}) a probability of classification is naturally produced, for other classifiers (\textit{svm}, \textit{trees}, \textit{forest}) this probability can be derived using additional methods implemented within the \texttt{sklearn} library.

For each random seed we determine the best hyper-parameters for each ML algorithm.
We train $30$ models for each ML algorithm representing different random combinations of hyper-parameters.
For \textit{logit} we pick the inverse of regularization strength from an uniform distribution $\mathcal{U}(0.1, 10)$. For \textit{mlp}, we use a two layer network with the sizes $(F,P*F),(P*F,(P+1)*F),(P*F,1)$, where $F$ is the number of input features and $P$ is chosen randomly from an uniform distribution $\mathcal{U}(1,10)$. In addition we experimentally determined the batch size to be $64$, we update parameters using the stochastic gradient descent for $100$ epochs. The cost function for \textit{mlp} classification is binary cross entropy, with an $\mathcal{L}_2$ penalty on weights of $0.01$ to avoid over-fitting. For \textit{knn} the number of neighbors and the distance metrics are picked randomly between $(3,20)$  and between Minkowski, Euclidean and Manhattan. For the \textit{lsvm} the penalty C is drawn from an uniform distribution $\mathcal{U}(0.1, 10)$. For the \textit{rf} we randomly pick the number of estimators to be between $(10, 50)$, the maximum depth between $(5, 50)$ and the minimum number of samples per leaf between $(1, 10)$.
We select as the best model for each ML algorithm, the one having the highest AUCROC on the validation set.

\subsubsection{Software implementation details.}
The experiments are replicated $10$ times for different seeds to ensure robustness and reproducibility. The code is implemented in Python using libraries such as \texttt{pandas} and \texttt{sklearn-pandas} for data processing, \texttt{sklearn} and \texttt{pytorch} for machine learning, \texttt{numpy} and \texttt{scipy} for numerical processing.
This research complies with research reproducibility principles, and code is made publicly available as a part of a framework.\footnote{HUMAINT repository: \href{https://gitlab.com/HUMAINT/humaint-fatml}{https://gitlab.com/HUMAINT/humaint-fatml}.}

\subsection{Results}

\subsubsection{Fairness analysis using clustering.}\label{sssec:clustering}
Towards a fairness analysis encompassing a wide variety of metrics we look at how group-wise measures, ML models, and groups cluster. 
The $\arr{M}_p$ matrix for a protected feature $p$ (e.g., race) comprises all computed metrics across all groups and for all ML models. The hierarchical clustering of $\arr{M}_p$ on lines and columns gives important information on how metrics relate to each other and on how groups differ across all metrics. 

\begin{figure}[ht!]
\centering\includegraphics[scale = 0.3]{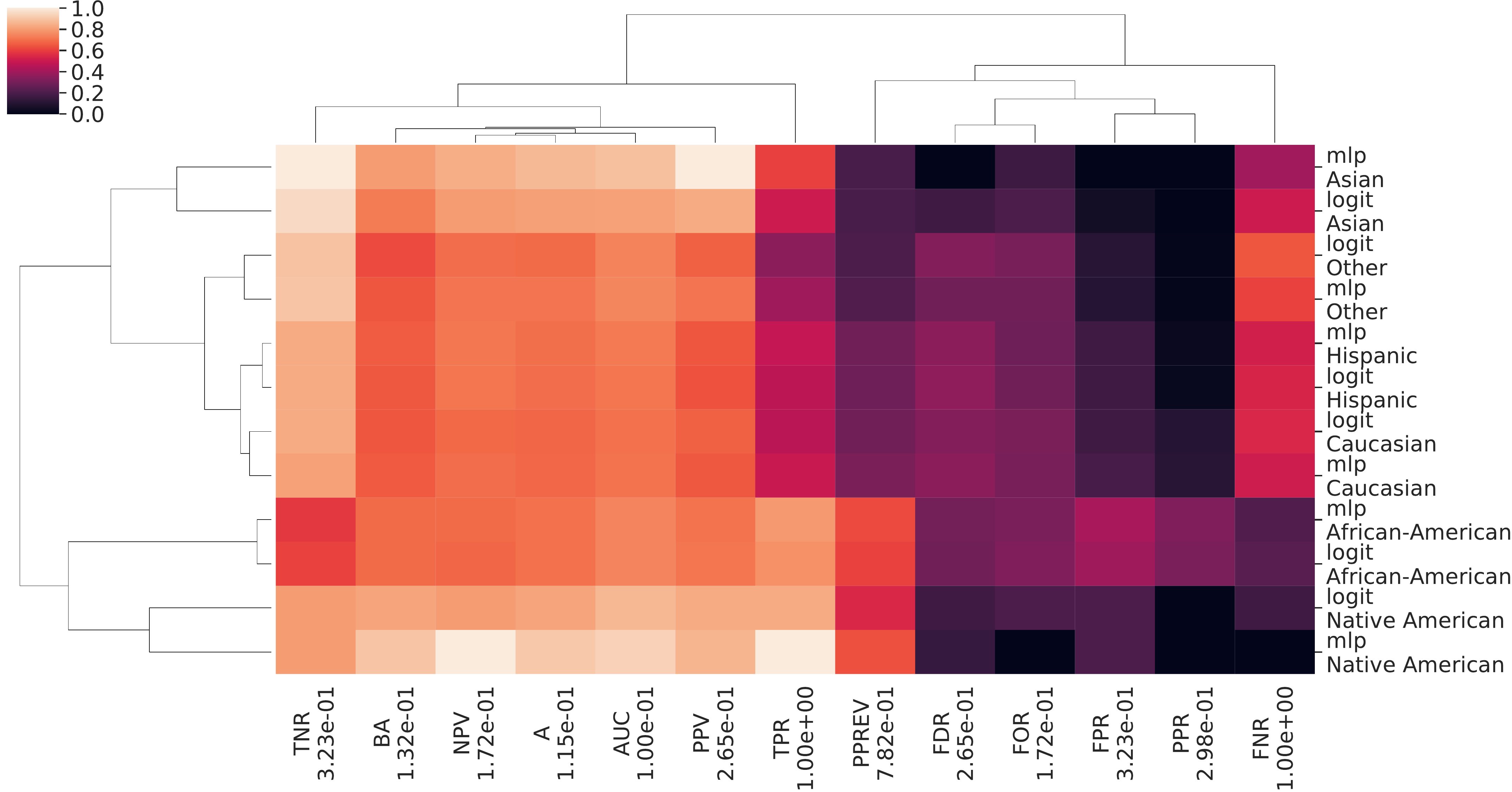}
\caption{The matrix $\arr{M}_p$ and the resulting hierarchical clustering for the metrics (columns), and the groups, ML models (lines) for dataset COMPAS and protected feature $'race'$. The metrics are presented along with the corresponding variances. }
\label{fig:cluster}
\end{figure}

Figure \ref{fig:cluster} shows an example of the matrix $\arr{M}_p$ and of resulting clusters for the COMPAS dataset and the protected feature $p=$\texttt{"race"} for a single seed. For brevity, we limit the number of ML models in the plot to the top two in terms of AUCROC: \textit{logit} and \textit{mlp}. 

The matrix $\arr{M}_p$ is presented as a table having the metrics on the columns, and the groups (Caucasian, African-American, Hispanic, Other, Native-American), ML models (\textit{logit} and \textit{mlp}) on the rows. We display the clustering information for the metrics above the columns and the clustering information of the groups and ML models on the left side of the lines. The clustering is computed according to the method in Section \ref{ssec:clusterigmet}.

First, we look at the clustering on the columns. We observe that the evaluation metrics cluster on two different groups. On one hand, we have $AUC, A, BA,$\linebreak $TPR, TNR, PPV, NPV$ and on the other hand, $FDR, FNR, FPR, FOR, PPR,$\linebreak $ PPREV$. We associate the first group with performance metrics and the second one with errors and prevalence metrics. The two main big clusters observed in Figure \ref{fig:cluster} is in line with the correlation between the metrics observed in \cite{friedler2019comparative}. 

We display the variance of each metric near the corresponding label. The variance and the clustering information can be used to choose a set of metrics to optimize for, e.g. the metrics with the highest variance from two separate clusters. Note that the variance is equal for complementary metrics.
There is less variance for accuracy metrics $AUC, A, BA$ which are closely related. Since the ML models are trained to optimize predictive performance, this finding is in line with the trade-off between accuracy and fairness in the literature. 
Furthermore, we observe that $TPR, FNR, PPREV$ have the largest variance and the largest disparity between groups occurs at these metrics. In addition, $TPR$ and $FNR$ are by definition complementary and optimizing for one's parity means that the other is also optimized. Furthermore, $TPR$, $FNR$, $PPREV$ do not cluster closely to other metrics.

Second, we look at the clustering on the rows to assess for disparity between different groups and methods. 
With respect to the first cluster of metrics, African-Americans have lower $TNR$ for \textit{logit} and \textit{mlp}, meaning that they are less likely to be correctly labeled as non-recidivists, and higher $TPR$, meaning that they are more likely to be correctly label as recidivists.
With respect to the second cluster of metrics, a higher proportion of African Americans and Native Americans are classified as recidivists when compared to other groups (higher $PPREV$), and the ML models are less likely to wrongly classify them as recidivists.
These disparities yielded by ML models on the COMPAS dataset were widely presented in the FAT-ML literature, however using solely one or two fairness metrics.
%

\subsubsection{Robustness testing across datasets.}\label{sssec:correlation}

\begin{figure}[ht!]
\centering\includegraphics[scale = 0.3]{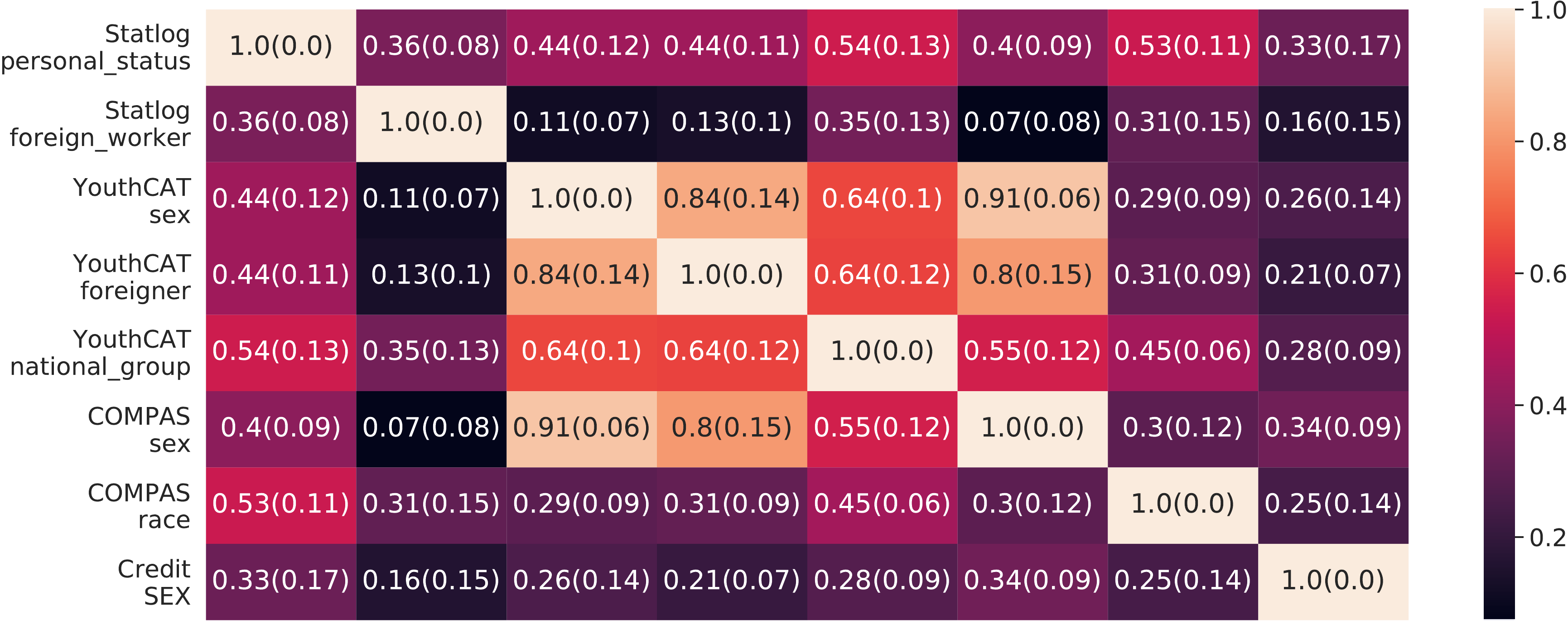}
\caption{The means and standard deviations of the correlations of vectors $\vektor{d}^I_p$ for all datasets and protected features. Columns are in the same ordering as rows.}
\label{fig:correlation}
\end{figure}

Considering the clusters seen in Section~\ref{sssec:clustering} for the matrix $\arr{M}_p$, we perform a PCA decomposition of matrix $\arr{M}_p$ for all datasets and seeds to determine the number of components and the explained variance. 
The means for the percentage of explained variances of the first two components are $0.69$ and $0.26$.
In this case, across all machine learning methods and datasets, the first two components explain $0.95$ of variance between the groups in the dataset across for the chosen classification metrics. 
This finding is consistent with the two main clusters observed in Figure \ref{fig:cluster}. Note that similar clusters were observed across other datasets and protected features, a fact which is confirmed by high correlations coefficients in Figure \ref{fig:correlation}. With respect to the correlation of metrics observed in Friedler et al.~\cite{friedler2019comparative}, its robustness is not tested across multiple datasets.

%
We want to assess whether the clusters observed in Figure~\ref{fig:cluster} are consistent across other datasets and protected features.
To do so, we compute correlation coefficients between all distance vectors $\vektor{d}^I_p$ on all protected features and datasets.
Means and standard deviations of the correlation coefficients across all seeds are reported in Figure~\ref{fig:correlation}.

We observe that the distance vectors for YouthCAT and COMPAS across the selected protected features are highly correlated. The two datasets point out to a similar scenario, criminal recidivism. However,  in the case of COMPAS, race yields less correlation with the national group or foreigner. In fact, the YouthCAT does not implicitly hold race as a feature, although national groups may encode different races. This points out to the fact that the categories considered in the dataset may yield different results in terms of group fairness \cite{benthall2019racial}.

Despite the fact that each dataset and protected feature yields two or three clusters, the way the metrics are distributed between the clusters is scenario dependent. Hence, except for COMPAS and YouthCAT metrics obtained for other datasets do not have a high correlation.


The particular case of Statlog shows that distance vectors are poorly correlated with the ones from the other three datasets.
This can be explained by the fact that Statlog presents a different type of problem, as the decision making is assistive, unlike YouthCAT, COMPAS, and Credit for which the decision is punitive.
This simple fact changes the meaning of the labels and the meaning of the metrics.
Note that in the comparative analysis of fairness-enhancing methods \cite{friedler2019comparative} the datasets for which the decision is punitive are analyzed separately from the assistive ones.

\subsubsection{Multi-metric fairness visualization using PCA decomposition.}\label{sssec:pcavis}
We aim at visualizing the distance between the ML methods and groups across all metrics in a lower dimension. We use the method described in Section \ref{ssec:pcamet}. 
The plots are centered at the reference group, considered here the largest among all groups. 

\begin{figure}[ht!]
\centering\includegraphics[scale = 0.6]{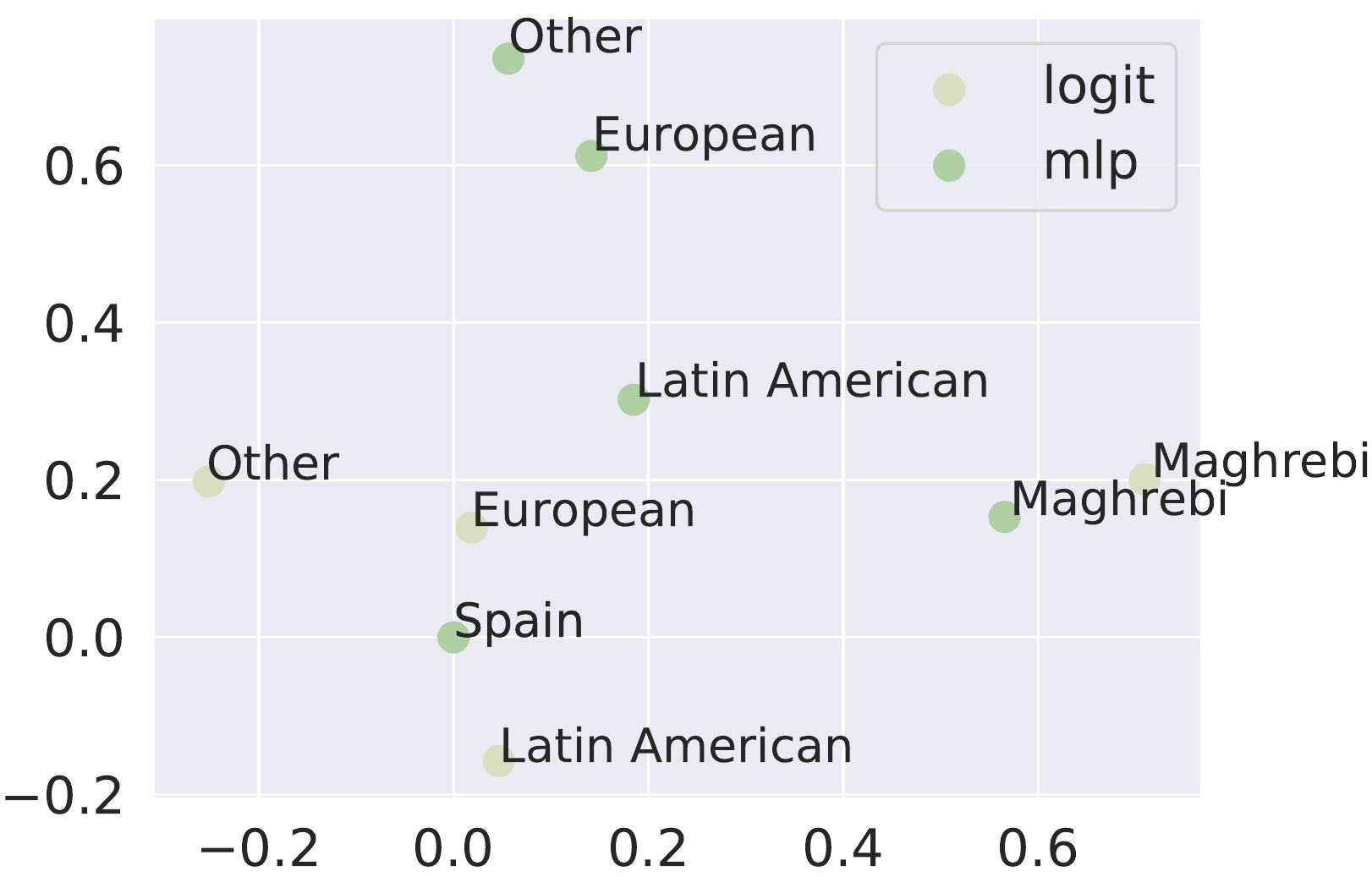}
\caption{The PCA $\hat{\arr{M}}_{p,l}$ for $l=\{\textit{logit},\textit{mlp}\}$ and $p=$\texttt{"national\_group"} for YouthCAT. The fraction of explained variance for the components are: $0.60,0.31,0.09$.}
\label{fig:pca}
\end{figure}

Figure \ref{fig:pca} shows the data points corresponding to the PCA matrices $\hat{\arr{M}}_{p,l}$ aligned between $L$ ML models for the dataset YouthCAT and the protected feature \texttt{"national\_group"}. For brevity, we limit the number of ML models to the top two ($L=2$) in terms of AUCROC: \textit{logit} and \textit{mlp}.

The groups in Figure \ref{fig:pca} for the dataset YouthCAT are Spanish, European, Latin American, Maghrebi, and Other. The reference group on which the plot is centered is Spanish. We observe that points corresponding to Maghrebi are far from the reference group and from all the other groups on the axis determined by the first component. Similarly, the Europeans and Others are far from the reference group on the axis determined by the second component.

While PCA axes do not hold any specific meaning in contrast to the classification metrics, they can give the magnitude of disparity, which is not easily accessible through the clustering in Figure \ref{fig:cluster}. However, the axis represent a linear combination of these metrics which were proven to be highly correlated in Section~\ref{sssec:clustering} and in Friedler et al.~\cite{friedler2019comparative}. The coefficients of this linear combination can be easily obtained from the eigen vectors. 


\section{Conclusions}\label{sec:conclusions}

In this paper we propose a reproducible and open methodology to visualize and study group fairness in data-driven decision making, beyond the limitations of an analysis relying on a very limited set of metrics. 
The context in which this framework is developed is characterized by various facets and definitions of fairness metrics \cite{romei2014multidisciplinary,zliobaite2017measuring,Barocas2016} and impossibility theorems \cite{kleinberg2016inherent,Chouldechova2017}, which prove that it is impossible to optimize for different fairness metrics. 
To that extent, we discover that fairness measures are highly correlated and it is convenient to visualize and assess fairness in two or three orthogonal dimensions.
Moreover, our experiments prove that the classification metrics group into two or three clusters. The resulting clusters do not generalize over the analyzed datasets and are dependent on each scenario.  

A two-dimensional reduction gives the possibility to compare different ML models in terms of fairness, to identify the groups affected by disparate impact. However, in this representation the axes do not hold any specific meaning and it is difficult to claim that a group is discriminated.
The authors recommend that the PCA analysis in Section~\ref{sssec:pcavis} is used in conjunction with the clustering plot in Section~\ref{sssec:clustering}.
While the former is useful to compare ML models and to have an initial measure of disparity, the latter offers information on what metrics are problematic for each groups and how these metrics are related.

\subsubsection{Limitations.}
The present study does not consider a comparison with decision-making systems which do not rely on ML, such as structured professional judgments, like \texttt{SAVRY}~\cite{cejfe_2017_savry}, which has been applied to the YouthCAT dataset. Neither does it conduct a comparison between fairness-enhancing methods. The results are reported for a set of machine learning methods and clusters of metrics. The variance of PCA components can change when including different systems in the evaluation. 

\subsubsection{Future work.}
The open source framework allows for the current methodology to be applied to any binary decision making dataset. We plan on extending the current study to include more datasets and a comparison with fairness-enhancing methods.

\bibliographystyle{splncs04}
\bibliography{biblio.bib}

\begin{thebibliography}{10}
\providecommand{\url}[1]{\texttt{#1}}
\providecommand{\urlprefix}{URL }
\providecommand{\doi}[1]{https://doi.org/#1}

\bibitem{adler2018auditing}
Adler, P., Falk, C., Friedler, S.A., Nix, T., Rybeck, G., Scheidegger, C.,
  Smith, B., Venkatasubramanian, S.: Auditing black-box models for indirect
  influence. Knowledge and Information Systems  \textbf{54}(1),  95--122 (2018)

\bibitem{agarwal2018reductions}
Agarwal, A., Beygelzimer, A., Dud{\'\i}k, M., Langford, J., Wallach, H.: A
  reductions approach to fair classification. arXiv preprint arXiv:1803.02453
  (2018)

\bibitem{Barocas_mimeo}
Barocas, S., Hardt, M., Narayanan, A.: Fairness and Machine Learning.
  fairmlbook.org (2018), \url{http://www.fairmlbook.org}

\bibitem{Barocas2016}
Barocas, S., Selbst, A.: {Big Data ' s Disparate Impact}. California law review
   \textbf{104}(1),  671--729 (2016). \doi{http://dx.doi.org/10.15779/Z38BG31},
  \url{https://ssrn.com/abstract=2477899}

\bibitem{benthall2019racial}
Benthall, S., Haynes, B.D.: Racial categories in machine learning. In:
  Proceedings of the Conference on Fairness, Accountability, and Transparency.
  pp. 289--298 (2019)

\bibitem{berk2017convex}
Berk, R., Heidari, H., Jabbari, S., Joseph, M., Kearns, M., Morgenstern, J.,
  Neel, S., Roth, A.: A convex framework for fair regression. arXiv preprint
  arXiv:1706.02409  (2017)

\bibitem{cejfe_2017_savry}
Blanch, M., Capdevila, M., Ferrer, M., Framis, B., Ruiz, U., Mora, J., Batlle,
  A., L\'{o}pez, B.: La reincid\`{e}ncia en la just\'{i}cia de menors. CEJFE
  (2017)

\bibitem{brennan2013emergence}
Brennan, T., Oliver, W.: The emergence of machine learning techniques in
  criminology: Implications of complexity in our data and in research
  questions. Criminology \& Public Policy  \textbf{12}(3),  551--562 (2013)

\bibitem{calmon2017optimized}
Calmon, F., Wei, D., Vinzamuri, B., Ramamurthy, K.N., Varshney, K.R.: Optimized
  pre-processing for discrimination prevention. In: Advances in Neural
  Information Processing Systems. pp. 3992--4001 (2017)

\bibitem{Chouldechova2017}
Chouldechova, A.: Fair prediction with disparate impact: A study of bias in
  recidivism prediction instruments. Big data  \textbf{5}(2),  153--163 (2017)

\bibitem{Corbett2017}
Corbett-Davies, S., Pierson, E., Feller, A., Goel, S., Huq, A.: Algorithmic
  decision making and the cost of fairness. In: Proceedings of the 23rd ACM
  SIGKDD International Conference on Knowledge Discovery and Data Mining. pp.
  797--806. ACM (2017)

\bibitem{day1984efficient}
Day, W.H., Edelsbrunner, H.: Efficient algorithms for agglomerative
  hierarchical clustering methods. Journal of classification  \textbf{1}(1),
  7--24 (1984)

\bibitem{dwork2012fairness}
Dwork, C., Hardt, M., Pitassi, T., Reingold, O., Zemel, R.: Fairness through
  awareness. In: Proceedings of the 3rd innovations in theoretical computer
  science conference. pp. 214--226. ACM (2012)

\bibitem{fish2016confidence}
Fish, B., Kun, J., Lelkes, {\'A}.D.: A confidence-based approach for balancing
  fairness and accuracy. In: Proceedings of the 2016 SIAM International
  Conference on Data Mining. pp. 144--152. SIAM (2016)

\bibitem{friedler2019comparative}
Friedler, S.A., Scheidegger, C., Venkatasubramanian, S., Choudhary, S.,
  Hamilton, E.P., Roth, D.: A comparative study of fairness-enhancing
  interventions in machine learning. In: Proceedings of the Conference on
  Fairness, Accountability, and Transparency. pp. 329--338. ACM (2019)

\bibitem{galhotra2017fairness}
Galhotra, S., Brun, Y., Meliou, A.: Fairness testing: testing software for
  discrimination. In: Proceedings of the 2017 11th Joint Meeting on Foundations
  of Software Engineering. pp. 498--510. ACM (2017)

\bibitem{hajian2016algorithmic}
Hajian, S., Bonchi, F., Castillo, C.: Algorithmic bias: From discrimination
  discovery to fairness-aware data mining. In: Tutorial at the 22nd {ACM
  SIGKDD} International Conference on Knowledge Discovery and Data Mining. pp.
  2125--2126. ACM (2016)

\bibitem{Hardt2016}
Hardt, M., Price, E., Srebro, N., et~al.: Equality of opportunity in supervised
  learning. In: Advances in neural information processing systems. pp.
  3315--3323 (2016)

\bibitem{kamiran2012data}
Kamiran, F., Calders, T.: Data preprocessing techniques for classification
  without discrimination. Knowledge and Information Systems  \textbf{33}(1),
  1--33 (2012)

\bibitem{kleinberg2016inherent}
Kleinberg, J., Mullainathan, S., Raghavan, M.: Inherent trade-offs in the fair
  determination of risk scores. In: Proc. of Innovations in Theoretical
  Computer Science ({ITCS}) (2017)

\bibitem{komiyama2018nonconvex}
Komiyama, J., Takeda, A., Honda, J., Shimao, H.: Nonconvex optimization for
  regression with fairness constraints. In: International Conference on Machine
  Learning. pp. 2742--2751 (2018)

\bibitem{lipton2018does}
Lipton, Z., McAuley, J., Chouldechova, A.: Does mitigating ml's impact
  disparity require treatment disparity? In: Advances in Neural Information
  Processing Systems. pp. 8136--8146 (2018)

\bibitem{liu2018delayed}
Liu, L.T., Dean, S., Rolf, E., Simchowitz, M., Hardt, M.: Delayed impact of
  fair machine learning. arXiv preprint arXiv:1803.04383  (2018)

\bibitem{menon2018cost}
Menon, A.K., Williamson, R.C.: The cost of fairness in binary classification.
  In: Conference on Fairness, Accountability and Transparency. pp. 107--118
  (2018)

\bibitem{narayanan2018translation}
Narayanan, A.: Translation tutorial: 21 fairness definitions and their
  politics. In: Proc. {FAT} Confernece on Fairness, Accountability, and
  Transparency (2018)

\bibitem{pedreschi2012study}
Pedreschi, D., Ruggieri, S., Turini, F.: A study of top-k measures for
  discrimination discovery. In: Proceedings of the 27th Annual ACM Symposium on
  Applied Computing. pp. 126--131. ACM (2012)

\bibitem{pleiss2017fairness}
Pleiss, G., Raghavan, M., Wu, F., Kleinberg, J., Weinberger, K.Q.: On fairness
  and calibration. In: Advances in Neural Information Processing Systems. pp.
  5680--5689 (2017)

\bibitem{robert2014machine}
Robert, C.: Machine learning, a probabilistic perspective (2014)

\bibitem{romei2014multidisciplinary}
Romei, A., Ruggieri, S.: A multidisciplinary survey on discrimination analysis.
  The Knowledge Engineering Review  \textbf{29}(5),  582--638 (2014)

\bibitem{saleiro2018aequitas}
Saleiro, P., Kuester, B., Stevens, A., Anisfeld, A., Hinkson, L., London, J.,
  Ghani, R.: Aequitas: A bias and fairness audit toolkit. arXiv preprint
  arXiv:1811.05577  (2018)

\bibitem{Miron2019}
Tolan, S., Miron, M., G\'{o}mez, E., Castillo, C.: Why machine learning leads
  to unfairness in juvenile justice: Evidence from catalonia. ICAIL '19:
  International Conference on Artificial Intelligence and Law (submitted)
  (2019)

\bibitem{yeh2009comparisons}
Yeh, I.C., Lien, C.h.: The comparisons of data mining techniques for the
  predictive accuracy of probability of default of credit card clients. Expert
  Systems with Applications  \textbf{36}(2),  2473--2480 (2009)

\bibitem{Zafar2017fairness}
Zafar, M.B., Valera, I., Rogriguez, M.G., Gummadi, K.P.: Fairness constraints:
  Mechanisms for fair classification. In: Artificial Intelligence and
  Statistics. pp. 962--970 (2017)

\bibitem{zemel2013learning}
Zemel, R., Wu, Y., Swersky, K., Pitassi, T., Dwork, C.: Learning fair
  representations. In: International Conference on Machine Learning. pp.
  325--333 (2013)

\bibitem{zliobaite2017measuring}
{\v{Z}}liobait{\.e}, I.: Measuring discrimination in algorithmic decision
  making. Data Mining and Knowledge Discovery  \textbf{31}(4),  1060--1089
  (2017)

\end{thebibliography}

\end{document}


\appendix

\begin{sidewaysfigure}
\centering\includegraphics[width=0.99\textheight]{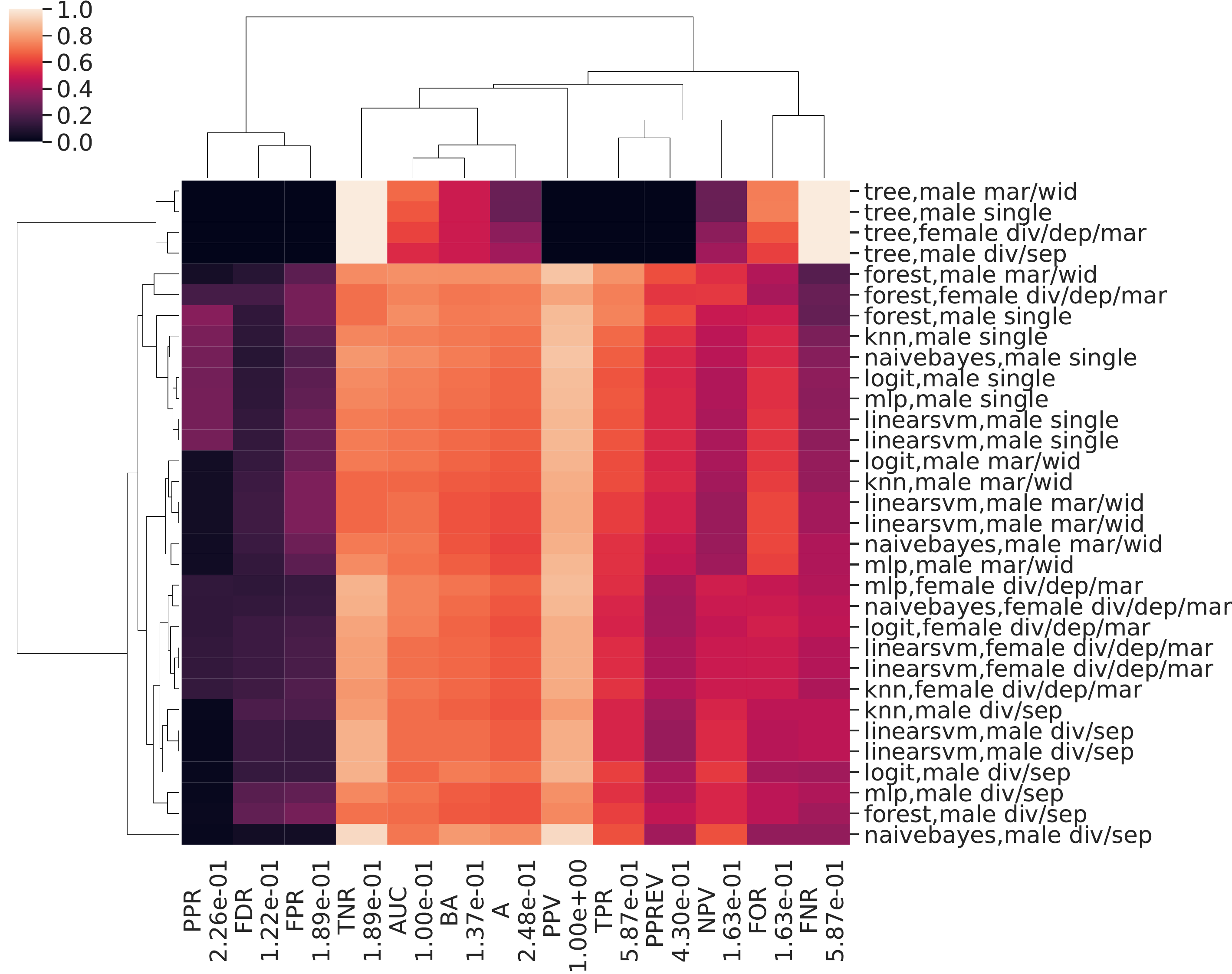}
\caption{The matrix $\arr{M}_p$ and the resulting hierarchical clustering for the metrics (columns), and the groups, ML models (lines) for dataset Statlog and protected feature $'personal\_status'$. The metrics are presented along with the corresponding variances. }
\label{fig:cluster}
\end{sidewaysfigure}

\begin{sidewaysfigure}
\centering\includegraphics[width=0.99\textheight]{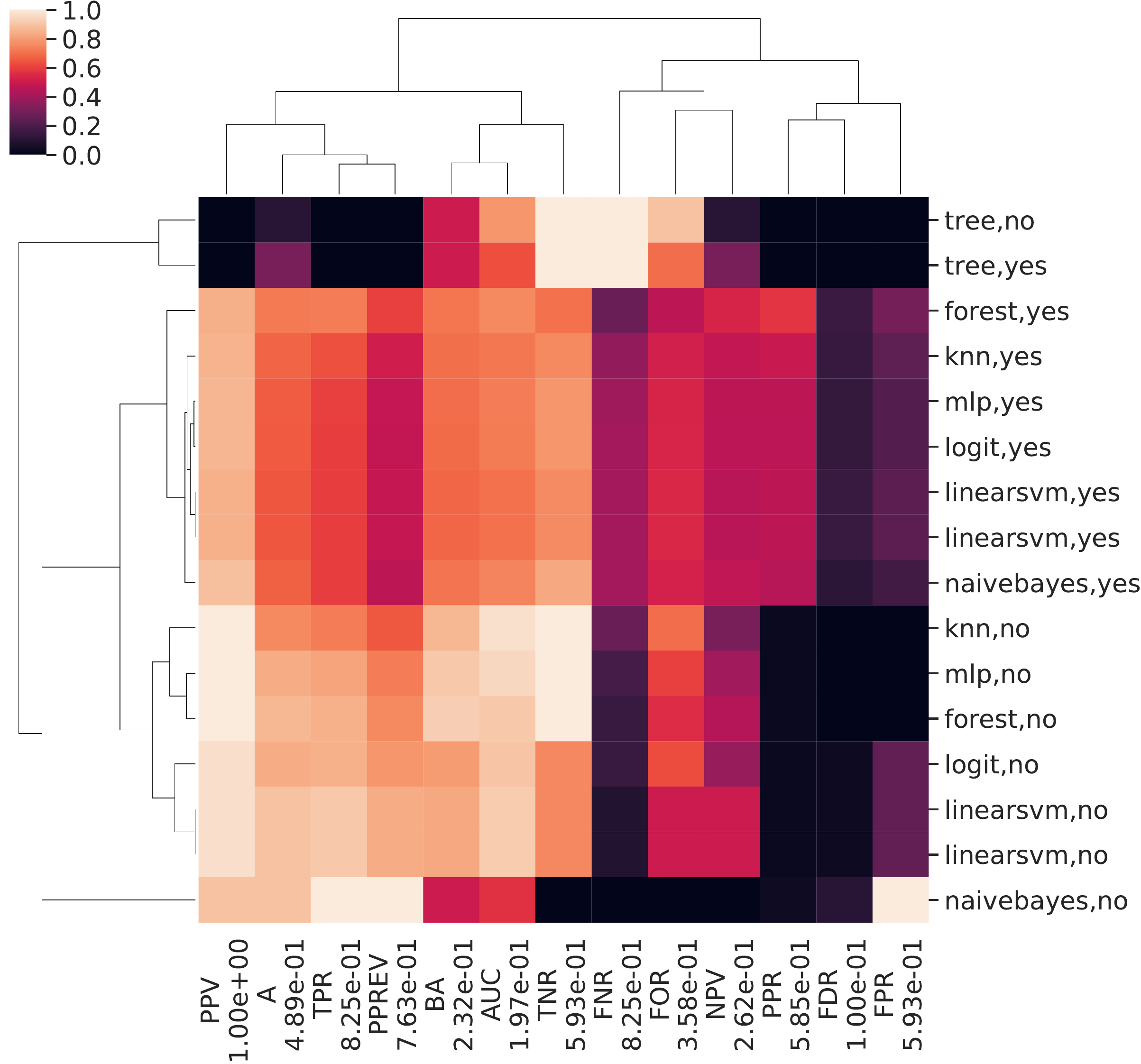}
\caption{The matrix $\arr{M}_p$ and the resulting hierarchical clustering for the metrics (columns), and the groups, ML models (lines) for dataset Statlog and protected feature $'foreigner\_worker'$. The metrics are presented along with the corresponding variances. }
\label{fig:cluster}
\end{sidewaysfigure}

\begin{sidewaysfigure}
\centering\includegraphics[width=0.99\textheight]{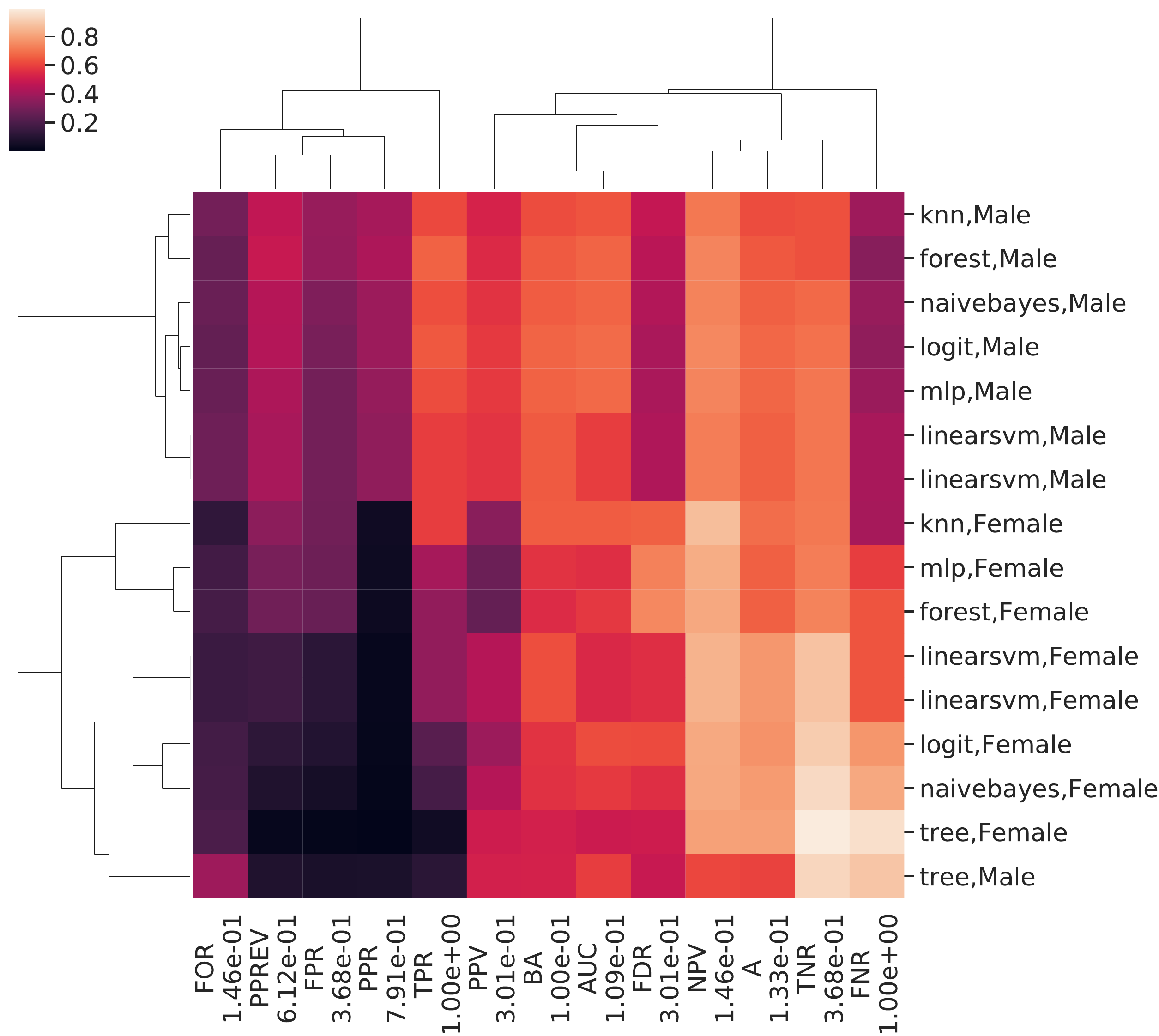}
\caption{The matrix $\arr{M}_p$ and the resulting hierarchical clustering for the metrics (columns), and the groups, ML models (lines) for dataset YouthCAT and protected feature $'sex'$. The metrics are presented along with the corresponding variances. }
\label{fig:cluster}
\end{sidewaysfigure}

\begin{sidewaysfigure}
\centering\includegraphics[width=0.99\textheight]{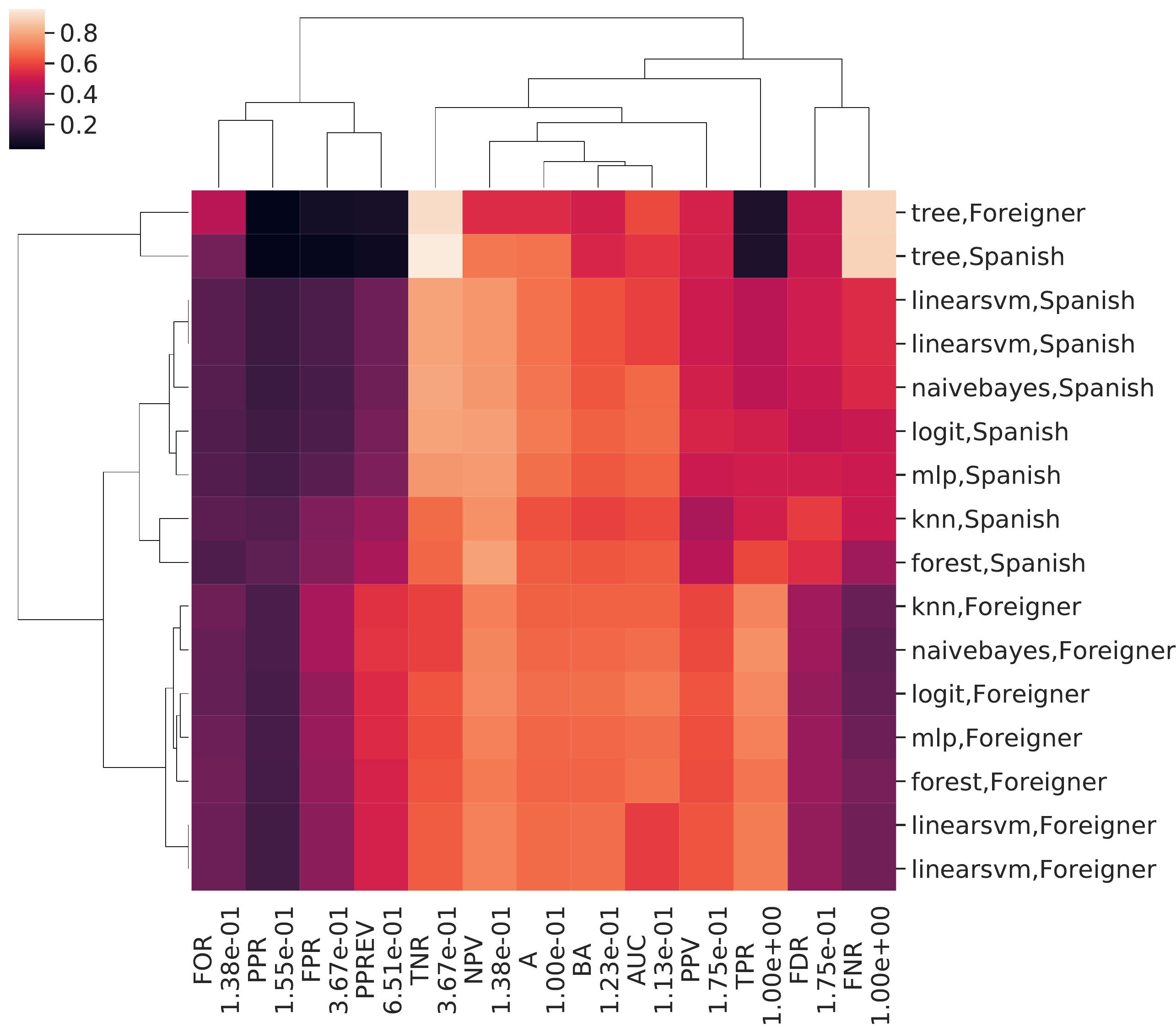}
\caption{The matrix $\arr{M}_p$ and the resulting hierarchical clustering for the metrics (columns), and the groups, ML models (lines) for dataset YouthCAT and protected feature $'foreigner'$. The metrics are presented along with the corresponding variances. }
\label{fig:cluster}
\end{sidewaysfigure}

\begin{sidewaysfigure}
\centering\includegraphics[width=0.99\textheight]{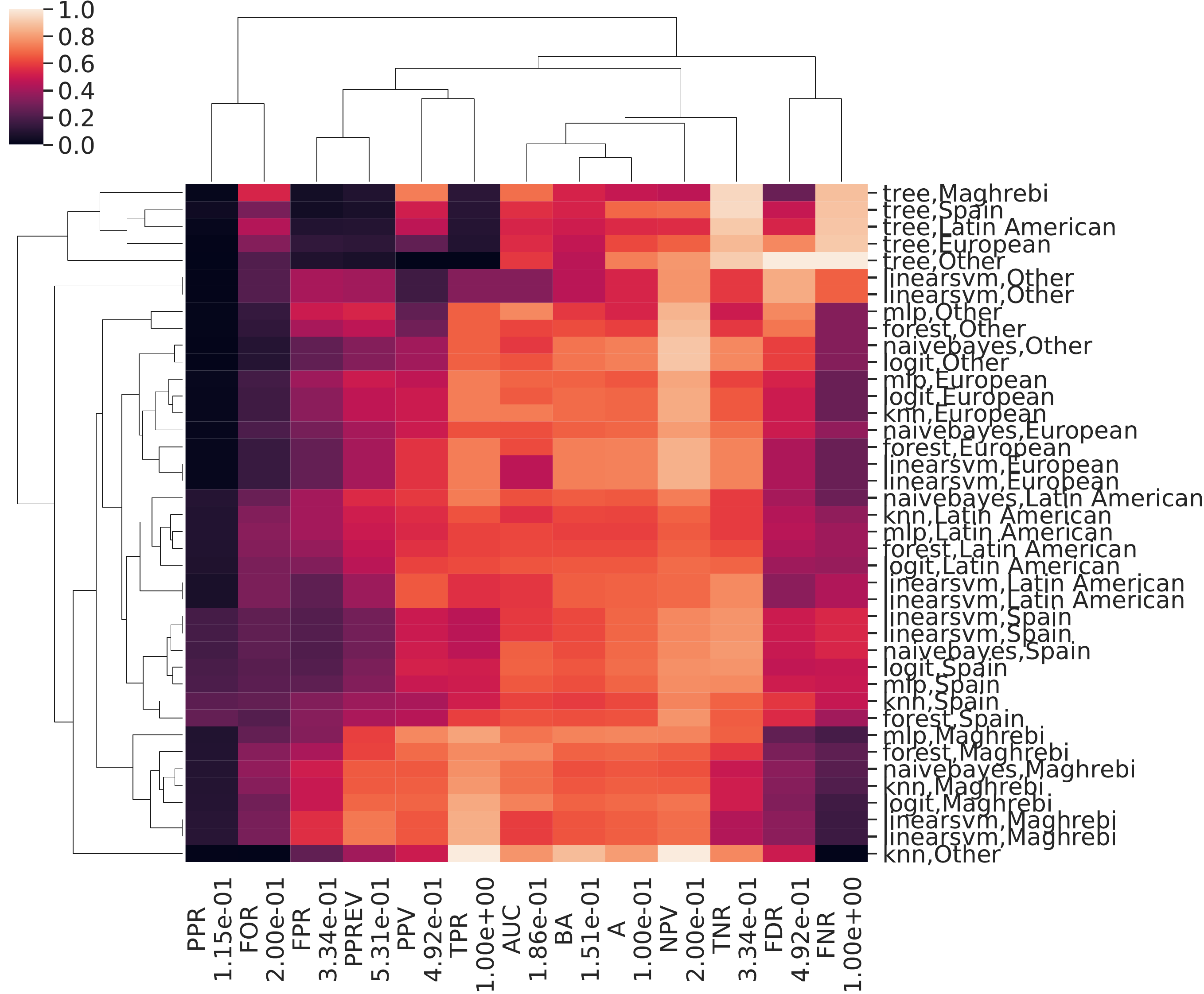}
\caption{The matrix $\arr{M}_p$ and the resulting hierarchical clustering for the metrics (columns), and the groups, ML models (lines) for dataset YouthCAT and protected feature $'national\_group'$. The metrics are presented along with the corresponding variances. }
\label{fig:cluster}
\end{sidewaysfigure}

\begin{sidewaysfigure}
\centering\includegraphics[width=0.99\textheight]{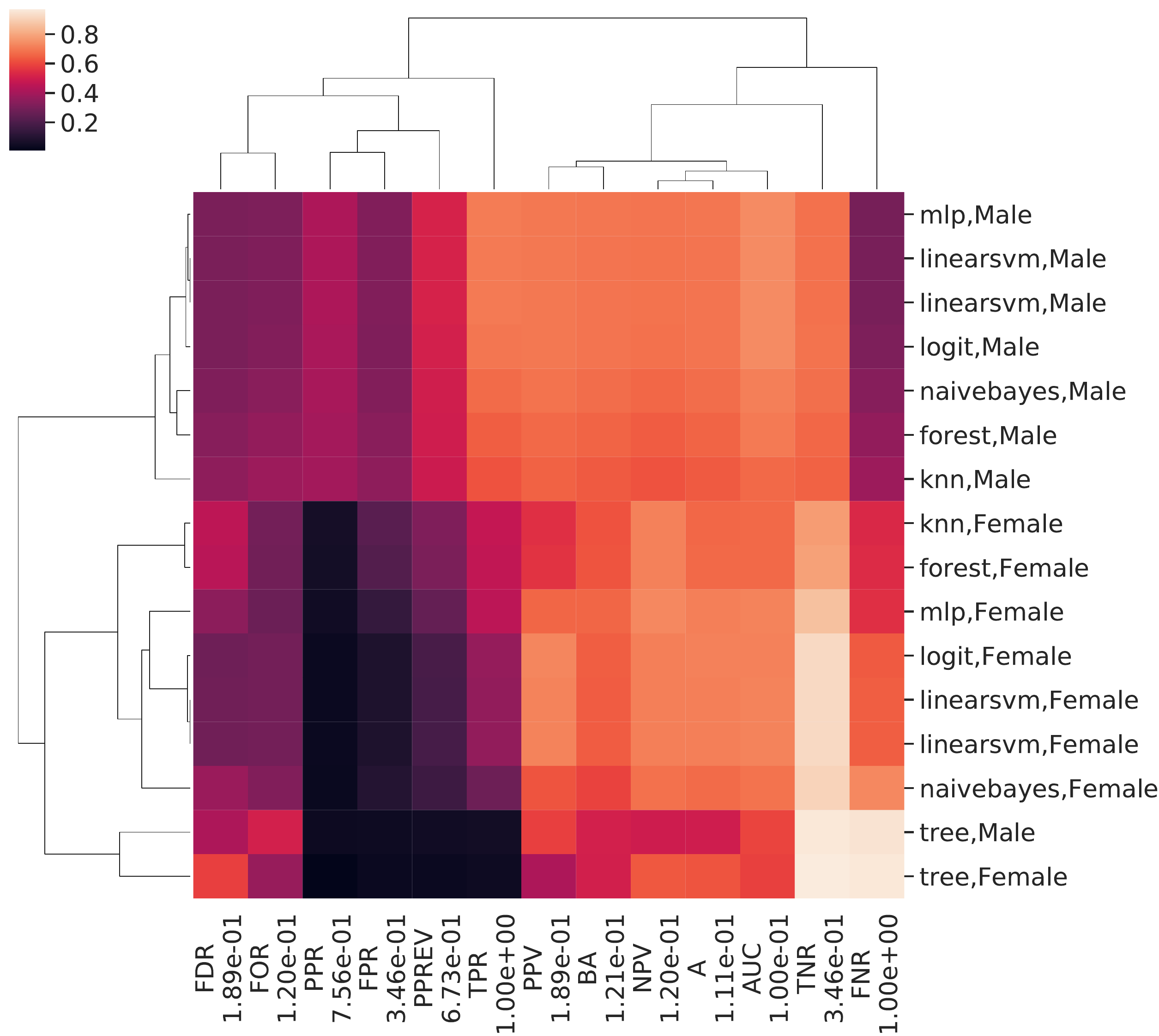}
\caption{The matrix $\arr{M}_p$ and the resulting hierarchical clustering for the metrics (columns), and the groups, ML models (lines) for dataset COMPAS and protected feature $'sex'$. The metrics are presented along with the corresponding variances. }
\label{fig:cluster}
\end{sidewaysfigure}

\begin{sidewaysfigure}
\centering\includegraphics[width=0.99\textheight]{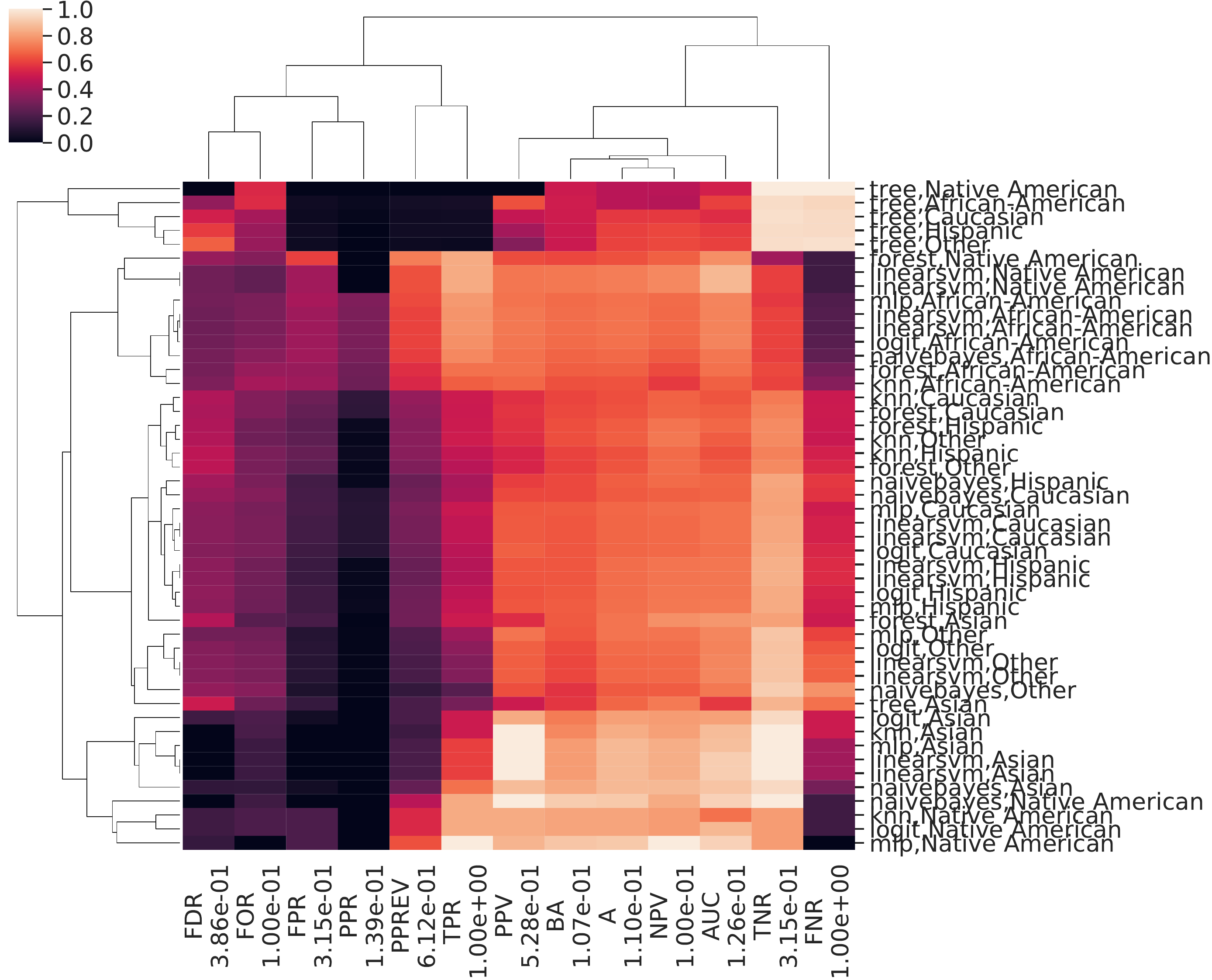}
\caption{The matrix $\arr{M}_p$ and the resulting hierarchical clustering for the metrics (columns), and the groups, ML models (lines) for dataset COMPAS and protected feature $'race'$. The metrics are presented along with the corresponding variances. }
\label{fig:cluster}
\end{sidewaysfigure}

\begin{sidewaysfigure}
\centering\includegraphics[width=0.99\textheight]{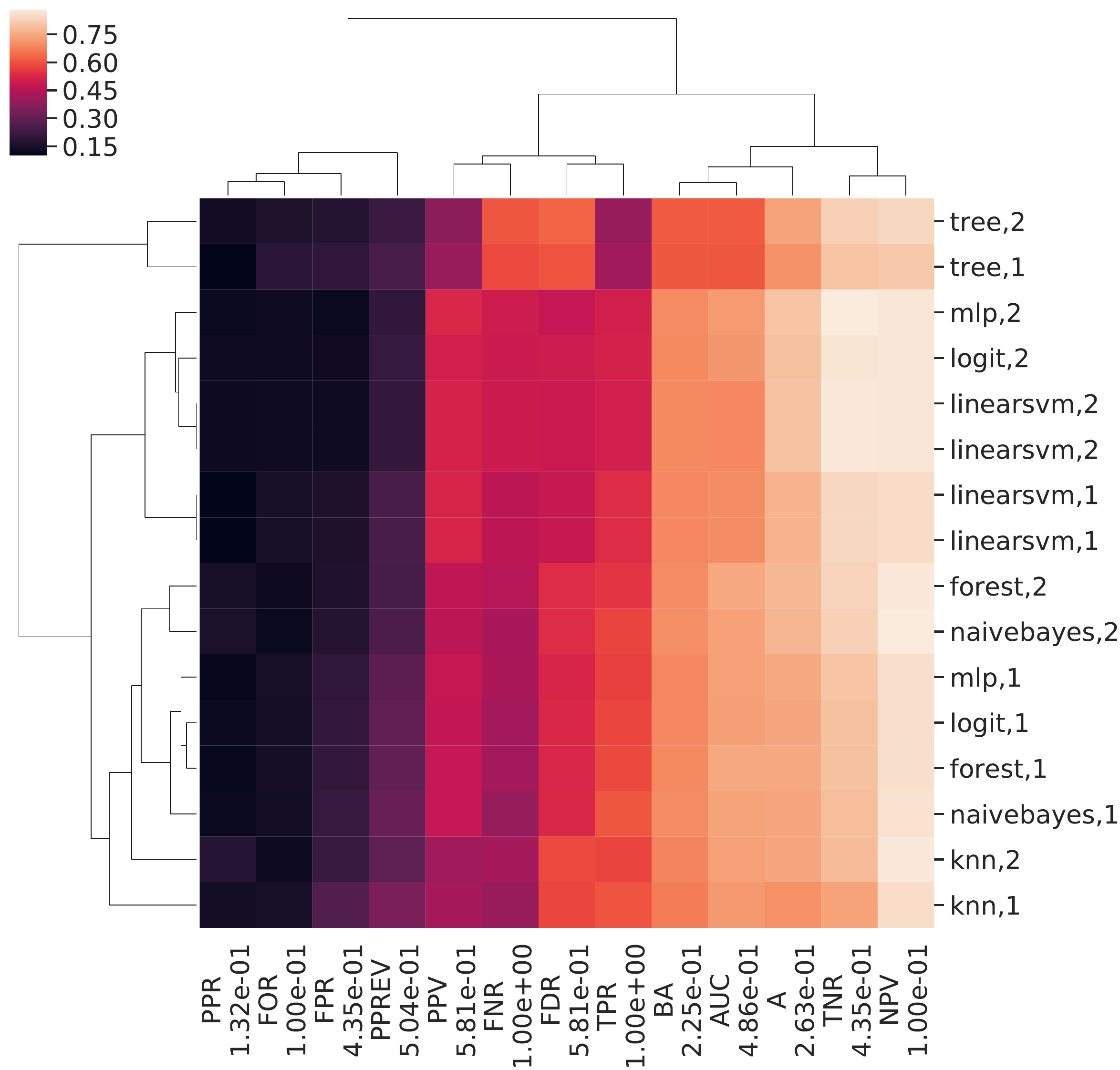}
\caption{The matrix $\arr{M}_p$ and the resulting hierarchical clustering for the metrics (columns), and the groups, ML models (lines) for dataset YouthCAT and protected feature $'sex'$. The metrics are presented along with the corresponding variances. }
\label{fig:cluster}
\end{sidewaysfigure}